\documentclass{article} 
\usepackage{iclr2025_conference,times}

\usepackage{amsmath,amsfonts,bm}

\def\eqref#1{equation~\ref{#1}}

\def\1{\bm{1}}

\DeclareMathAlphabet{\mathsfit}{\encodingdefault}{\sfdefault}{m}{sl}
\SetMathAlphabet{\mathsfit}{bold}{\encodingdefault}{\sfdefault}{bx}{n}

\usepackage{hyperref}
\usepackage{url}
\usepackage{graphicx}
\usepackage{amsmath}
\usepackage{booktabs}
\usepackage{multirow}
\usepackage{subcaption}
\usepackage{enumitem}
\usepackage{threeparttable}
\usepackage{wrapfig}

\newcommand{\piref}{\pi_\text{ref}}

\title{Enhancing Multimodal LLM for Detailed and Accurate Video Captioning using Multi-Round Preference Optimization}

\author{Changli Tang\textsuperscript{1}\thanks{Equal contribution},~ 
Yixuan Li\textsuperscript{1}\footnotemark[1],~
Yudong Yang\textsuperscript{1},~
Jimin Zhuang\textsuperscript{1},~ 
Guangzhi Sun\textsuperscript{2}, \\
\textbf{Wei Li\textsuperscript{3}},~
\textbf{Zejun Ma\textsuperscript{3}},~
\textbf{Chao Zhang\textsuperscript{1}\thanks{Corresponding author}} \\
Tsinghua University$^1$, University of Cambridge$^2$, ByteDance$^3$ \\ 
\texttt{\{tcl24, yixuan-l21\}@mails.tsinghua.edu.cn, cz277@tsinghua.edu.cn} \\
}

\iclrfinalcopy 
\begin{document}

\maketitle

\begin{abstract}
Videos contain a wealth of information, and generating detailed and accurate descriptions in natural language is a key aspect of video understanding. In this paper, we present video-SALMONN 2, an advanced audio-visual large language model (LLM) with low-rank adaptation (LoRA) designed for enhanced video (with paired audio) captioning through directed preference optimization (DPO). We propose new metrics to evaluate the completeness and accuracy of video descriptions, which are optimized using DPO. To further improve training, we introduce a novel multi-round DPO (mrDPO) approach, which involves periodically updating the DPO reference model, merging and re-initializing the LoRA module as a proxy for parameter updates after each training round (1,000 steps), and incorporating guidance from ground-truth video captions to stabilize the process. To address potential catastrophic forgetting of non-captioning abilities due to mrDPO, we propose rebirth tuning, which finetunes the pre-DPO LLM by using the captions generated by the mrDPO-trained model as supervised labels. Experiments show that mrDPO significantly enhances video-SALMONN 2's captioning accuracy, reducing global and local error rates by 40\% and 20\%, respectively, while decreasing the repetition rate by 35\%. The final video-SALMONN 2 model, with just 7 billion parameters, surpasses leading models such as GPT-4o and Gemini-1.5-Pro in video captioning tasks, while maintaining competitive performance to the state-of-the-art on widely used video question-answering benchmark among models of similar size. Upon acceptance, we will release the code, model checkpoints, and training and test data. Demos are available at \href{https://video-salmonn-2.github.io}{https://video-salmonn-2.github.io}.

\end{abstract} 

\section{Introduction}
Large language models (LLMs) have exhibited outstanding capabilities in a wide range of natural language processing (NLP) tasks, and in some instances, have even approached human-level performance \citep{openai2024gpt4technicalreport, dubey2024llama3herdmodels, llama, glm, qwen}. LLMs' remarkable ability to understand, generate, and reason with text has sparked widespread interest among researchers, attracting both academia and industry to extend them to multimodal understanding and generation. 
To endow LLMs with multimodal understanding capability, recent studies adopted a paradigm of training modality adapters and aligners between multi-modal encoders and LLMs. This approach leverages world knowledge in the textual LLM to interpret diverse types of data perceived by multimodal encoders, enabling the generation of meaningful insights.
Over the past two years, many multimodal LLMs have emerged following this paradigm across different modalities. These include models for image and silent video understanding \citep{liu2024visual, liu2024improved, li2023blip, Qwen-VL, lin2023vila, chen2023internvl, lin2023video, chen2024sharegpt4video}, audio understanding \citep{wu2023decoder, tang2024salmonn, Qwen-Audio, Qwen2Audio, gong2024listen, gong-ltuas, tang24d_interspeech, zhengbat}, and audio-visual understanding \citep{geminiteam2024geminifamilyhighlycapable, damonlpsg2024videollama2, sun2024videosalmonn, fu2024vita, tang2024avicuna}.

Text descriptions of multimodal data are critical for building multimodal LLMs. This is because most contemporary multimodal LLMs treat multimodal captions as a cornerstone task during pre-training or supervised fine-tuning (SFT), to align the representation spaces of multimodal encoders with that of textual large language models, helping LLMs recognise and understand events in multimodal data. Thus, collecting high-quality text descriptions paired with multimodal data is crucial for constructing high-performance multimodal LLMs, which implies training the model with more detailed and less hallucinated labels aligned with the multimodal data that could enhance the LLM’s ability to perform multimodal understanding and reasoning. 
In video understanding, generating detailed and accurate captions is crucial but challenging, as videos contain rich content that encompasses not only spatial features within individual visual frames but also audio-visual events that unfold across multiple frames over time.
However, very few multimodal LLM-related works focus on improving the quality of video captions, due to the lack of quantitative metrics for evaluating video captions and the absence of training methods to enhance the completeness of these descriptions while reducing the risks of hallucination. Additionally, while audio is typically paired with video and provides crucial, complementary information to the visual content, most current visual LLMs lack audio-understanding abilities, leading to the omission of audio information in the generated captions.

In this paper, we introduce video-SALMONN 2, a multimodal LLM that supports both audio and visual inputs and primarily focuses on detailed and holistic audio-visual captioning. Building upon an already well-trained visual LLM, video-SALMONN 2 is further enhanced with auditory capabilities by training on audio-only data as well as videos with synchronized audio tracks. This enables the model to simultaneously ``see" and ``hear" the video, emulating the way humans perceive and interpret multimedia content.
To accurately assess the performance of the model, new metrics to evaluate captioning quality are proposed, which then serve as the objective to optimize during reinforcement learning (RL) based on direct preference optimization (DPO). A novel multi-round DPO (mrDPO) is proposed and performed based on the preferences guided by the metrics, followed by a novel rebirth tuning stage to avoid the degradation of the non-captioning abilities caused by the mrDPO. The rebirth tuning leverages the post-mrDPO model to revise the captions of the videos in the training set, and trains the model after audio modality alignment using supervised fine-tuning (SFT) with the revised training data. 
Experiments demonstrate that video-SALMONN 2 with 7 billion (B) parameters can generate complete and accurate video descriptions and even outperforms much larger commercial multimodal LLMs such as GPT-4o and Gemini-1.5-Pro, and it also maintains competitive performance to the state-of-the-art (SOTA) multimodal LLM of similar model size on the commonly used Video-MME \citep{fu2024video} video question-answering (QA) benchmark.



The main contributions of this work can be summarised as follows:
\begin{itemize}[itemsep=0pt, leftmargin=*]
    \item We develop video-SALMONN 2, a powerful audio-visual LLM that generates high-quality video captions, outperforming larger commercial models such as GPT-4o and Gemini-1.5 in terms of completeness and accuracy.
    \item We introduce an evaluation pipeline that computes the missing and hallucination rates of audio-visual events in video captions using text-based LLMs, breaking down the process into sub-tasks suited for current LLMs. Additionally, we provide a new benchmark for video captioning with a human-annotated test set.
    \item We propose the mrDPO approach to optimize multimodal LLMs for video captioning, incorporating periodic updates to the DPO reference model, merging and reinitializing the low-rank adaptation (LoRA) \citep{hu2022lora} module, and smoothing the training loss using SFT based on ground-truth captions. To our knowledge, this is the first work applying RL to audio-visual LLMs.
    \item We introduce rebirth tuning to ensure the resulting model maintains high performance in both captioning and non-captioning tasks. The mrDPO process, followed by rebirth tuning, can be iteratively applied to further enhance performance. 
\end{itemize}





\section{Related Work}
\subsection{Multimodal LLMs}
Following the paradigm of connecting multimodal encoders to LLMs using modality adapters, various models have been developed. For image-based LLMs, LLaVA \citep{liu2024visual, liu2024improved} applies instruction tuning \citep{weifinetuned} to enhance performance on zero-shot tasks. BLIP-2 \citep{li2023blip} uses Q-Former to link a frozen encoder with an LLM, while VILA \citep{lin2023vila} explores pre-training strategies, achieving impressive results in video QA. InternVL \citep{chen2023internvl} scales up the size of visual encoders for improved image representation. For silent video understanding, Video-LLaVA \citep{lin2023video} aligns both image and video adapters to learn unified representations. ShareGPT4Video \citep{chen2024sharegpt4video} uses GPT-4 to generate dense video captions, improving data quality, and LLaVA-Hound \citep{zhang2024directpreferenceoptimizationvideo} introduces DPO to enhance video LLMs' understanding capabilities.

In the realm of audio perception, SALMONN \citep{tang2024salmonn} uses a dual-encoder structure and can perform zero-shot audio reasoning tasks. LTU \citep{gong2024listen} and LTU-AS \citep{gong-ltuas} trained on a large audio question-answering dataset are able to answer open-ended questions about audio. Qwen-Audio \citep{Qwen-Audio} and Qwen2-Audio \citep{Qwen2Audio} are built on large amounts of audio data to achieve high performance on a wide range of carefully selected audio tasks. \citet{zhengbat} and \citet{tang24d_interspeech} extend the LLM to perceive spatial audio information obtained from microphone array recordings.

As the visual frame sequence is often paired with audio in real-world video recordings, some studies investigate understanding non-silent video. video-SALMONN \citep{sun2024videosalmonn} uses a multi-resolution causal Q-Former to understand audio and video simultaneously. The Google Gemini model achieves video understanding as a native multimodal LLM built upon text, audio, and visual tokens \citep{geminiteam2024geminifamilyhighlycapable}.  AVicuna \citep{tang2024avicuna} achieves audio-visual temporal understanding by introducing pseudo-untrimmed video data. Video-LLaMA \citep{damonlpsg2023videollama} and Video-LLaMA 2 \citep{damonlpsg2024videollama2} directly concatenating audio and visual tokens for joint audio and video understanding.

\subsection{RL for LLMs}
RL with human feedback (RLHF) \citep{ouyang2022training} is commonly used to enhance text-based LLMs, with early efforts applying PPO \citep{schulman2017proximalpolicyoptimizationalgorithms} alongside a reward model trained on human preference data. Building on this, DPO \citep{rafailov2024direct} proposes that the LLM itself can serve as a reward model, using paired preference data to optimize the model without the need for an external reward model. KTO \citep{ethayarajh2024kto} further eliminates the need for paired preference data. Expanding on this, RLAIF \citep{lee2023rlaif} takes a cost-efficient approach by utilizing feedback generated automatically by models, reducing reliance on human involvement.


\section{Methods}
\begin{figure}[ht]
    \centering
    \includegraphics[width=\linewidth]{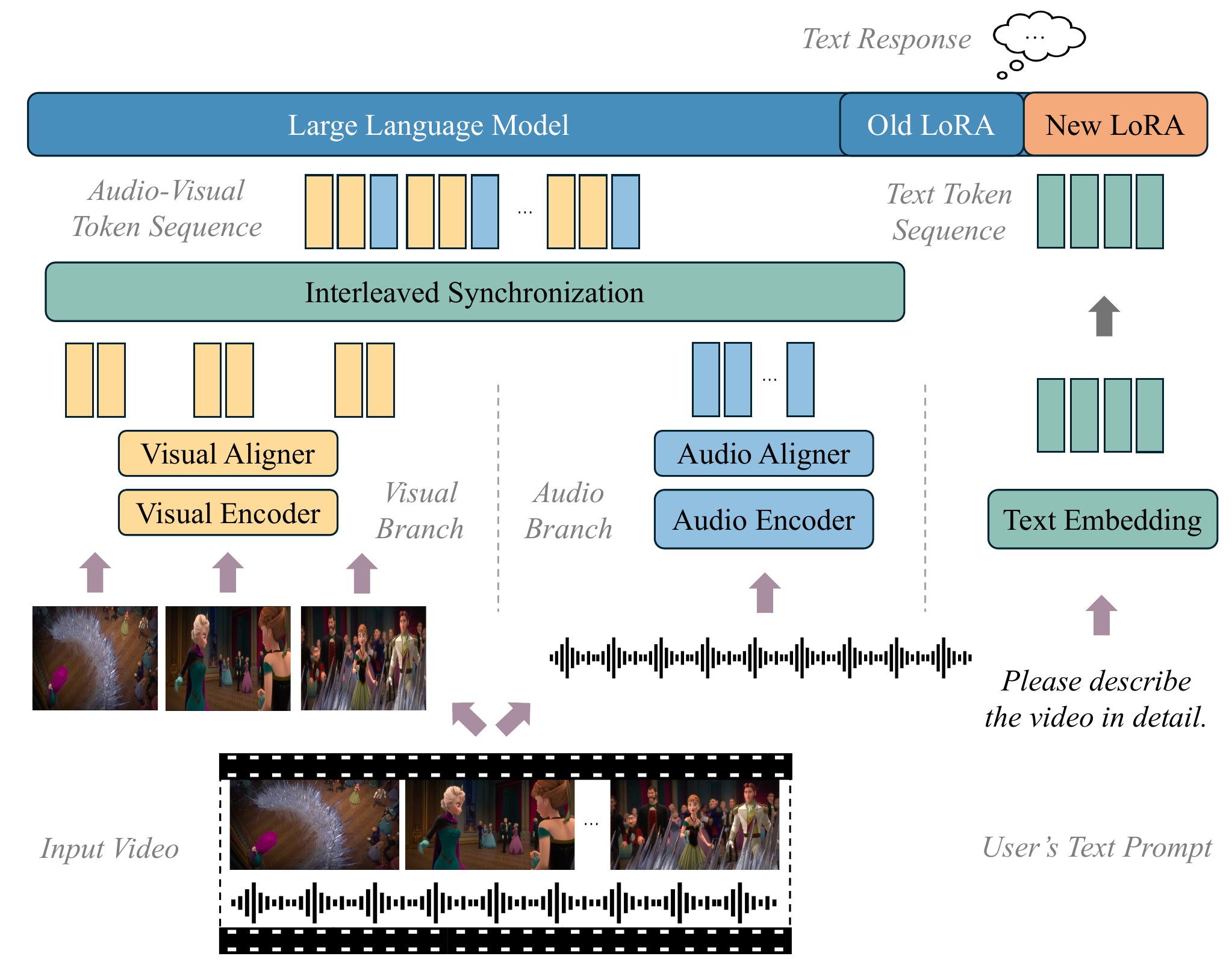}
    \caption{The overall structure of video-SALMONN 2. The input video is processed separately by the visual and audio branches, generating visual and audio tokens from the visual and audio frame sequences. Next, the visual and audio tokens are interleaved synchronously, and combined with the tokens of the text prompt to form the input to the LLM backbone.}
    \label{fig:avllava}
    \vspace{-0.5cm}
\end{figure}

\subsection{Model Architecture}
The overall architecture of our model is illustrated in Fig.~\ref{fig:avllava}. The paired sequences of audio and visual frames from each video are fed into the audio and visual encoders separately. Users can provide textual prompts to guide the model in performing specific tasks based on the video content.
This structure is implemented by incorporating a separate audio encoder branch to a pre-trained visual LLM, which enables the model to process and understand paired audio-visual sequences without degrading its visual performance.



In this structure, audio and visual tokens are computed independently in their respective branches. For the visual branch, the input visual frame sequence is first downsampled at a fixed frame rate of $\phi$ frame/second, and the total number of frames to sample is $n=\phi\,T$, where $T$ is the duration of the input video in seconds. Let $m$ be the maximum number of frames to sample based on the resource constraint. If $n>m$, the frame rate is further reduced to $\phi'=\lfloor m/T\rfloor$, resulting in $n=\phi'T\le m$. Let $\mathbf{I}_i$ be the $i$th sampled visual frame, and each visual frame in $\mathbf{I}_1,\mathbf{I}_2,\ldots,\mathbf{I}_n$ is transformed to visual tokens independently using a pre-trained visual encoder $\text{Encoder}_\text{Visual}$ followed by a visual modality aligner $\text{Aligner}_\text{Visual}$, as shown in Eqn.~(\ref{equ:video_br}): 
\begin{equation}
    \mathbf{H}^\text{Visual}_{i} = \text{Aligner}_\text{Visual}(\text{Encoder}_\text{Visual}(\mathbf{I}_i)),~~1\le i\le n,
    \label{equ:video_br}
\end{equation}
where $\mathbf{H_\text{V}}_{i}$ represents the visual tokens corresponding to $\mathbf{I}_i$.

The audio frame sequence $\mathbf{S}$ is fed into a pre-trained audio encoder $\text{Encoder}_\text{Audio}$. Since $\text{Encoder}_\text{Audio}$ may have a maximum processing duration $t_{\text{max}}$, the audio will be sliced into $l=\lceil{T}/{t_{\text{max}}\rceil}$ segments of $t_{\text{max}}$-length and processed separately by $\text{Encoder}_\text{Audio}$, as shown in Eqn.~(\ref{equ:audio_br1}):
\begin{equation}
    \mathbf{Z}^{\text{Audio}}_{j} = \text{Encoder}_\text{Audio}(\mathbf{S}_{(j-1)\times t_{\text{max}}:j\times t_{\text{max}}}),~~1\le j\le l,
    \label{equ:audio_br1}
\end{equation}
where $\mathbf{Z}^\text{Audio}_{j}$ is the audio feature vector output by the audio encoder of the $j$th audio segment.

As suggested by \citet{yu2024connect}, a segment-level positional embedding is added before the modality aligner to improve the performance of long-form audio. 
Denote $\mathbf{Z}^\text{Pos}_{j}$ as the segment-level position embedding matrix corresponding to the position $j$, $\text{Concat}(\cdot)$ as the concatenation operation along the time dimension, and $\text{Aligner}_{\text{Audio}}$ as the audio modality aligner. The audio token sequence $\mathbf{H}^{\text{Audio}}$ for the whole audio can be computed as Eqn.~(\ref{equ:audio_br21})--(\ref{equ:audio_br23}) shown:
\begin{align}
    \label{equ:audio_br21}
    \tilde{\mathbf{Z}}^\text{Audio}_{j} &= \mathbf{Z}^\text{Audio}_{j}+\mathbf{Z}^\text{Pos}_{j},~~1\le j\le l \\ 
    \tilde{\mathbf{Z}}^{\text{Audio}} &= \text{Concat}(\tilde{\mathbf{Z}}^\text{Audio}_{1},\tilde{\mathbf{Z}}^\text{Audio}_{2},\ldots,\tilde{\mathbf{Z}}^\text{Audio}_l) \\
    \mathbf{H}^{\text{Audio}} &= \text{Aligner}_{\text{Audio}}(\tilde{\mathbf{Z}}^\text{Audio}).
    \label{equ:audio_br23}
\end{align}
Next, the audio and visual tokens are interleaved chronologically to form the input audio-visual token sequence $\mathbf{H}$ fed into the LLM backbone, and $\mathbf{H}$ is obtained based on Eqn.~(\ref{equ:av1})--(\ref{equ:av2}) by
\begin{align}
    \label{equ:av1}
    \alpha_i &= l\cdot{i}/{n}, ~~1\leq i\leq n\\
    \mathbf{H}_i &= \text{Concat}(\mathbf{H}^\text{Visual}_i, \mathbf{H}^{\text{Audio}}_{\alpha_{i-1}:\alpha_{i}}),~~1\leq i\leq n \\
    \mathbf{H} &= \text{Concat}(\mathbf{H}_1,\mathbf{H}_2,\ldots,\mathbf{H}_{n}).
    \label{equ:av2}
\end{align}
Finally, the text-based backbone LLM is required to generate a text response $\mathbf{\hat{Y}}$ given the user's text prompt $\mathbf{P}$ and the audio-visual token sequence $\mathbf{H}$:
\begin{equation}
    \mathbf{\hat{Y}} = \arg\max\nolimits_\mathbf{Y}P(\mathbf{Y}|\mathbf{P}, \mathbf{H}).
\end{equation}


\subsection{Training Strategies}
To introduce audio perceptual capabilities to the visual LLM, we employ a multi-stage training approach that enables the model to fully utilize audio information for video understanding while maintaining its performance in processing visual data. Building on a well-trained visual LLM, the training proceeds through several stages: audio modality alignment, audio-visual SFT, RL based on the proposed mrDPO, and the newly introduced rebirth tuning.
For the pre-trained visual LLM that already understands video, with the LLM, visual aligner, and visual encoder well-initialized, both the LLM and video branch are kept frozen during all training stages. Similarly, the audio encoder parameters remain fixed, as they have already been trained on a large-scale audio dataset.

Audio modality alignment extends the visual LLM by adding a parallel audio branch, enabling auditory perception capabilities. During this stage, only the audio aligner is trained on a large audio dataset, while the rest of the model remains frozen to preserve its original visual understanding performance. Since the focus is exclusively on learning the audio branch, only audio data is needed for training.


After audio modality alignment, the backbone LLM can recognize both visual and audio tokens. However, due to the lack of training with paired audio and visual token sequences, the model is not yet capable of synchronizing and integrating audio-visual information for comprehensive video understanding. To address this, we conduct audio-visual SFT using supervised video data.
To improve the backbone LLM's ability to process audio-visual token sequences, LoRA \citep{hu2022lora} is applied and trained during this stage. Additionally, the audio aligner is trained to align the output of the audio encoder with the input representation space of the LLM, making it easier for the backbone LLM to interpret audio tokens.



\begin{figure}[ht]
    \centering
    \includegraphics[width=\linewidth]{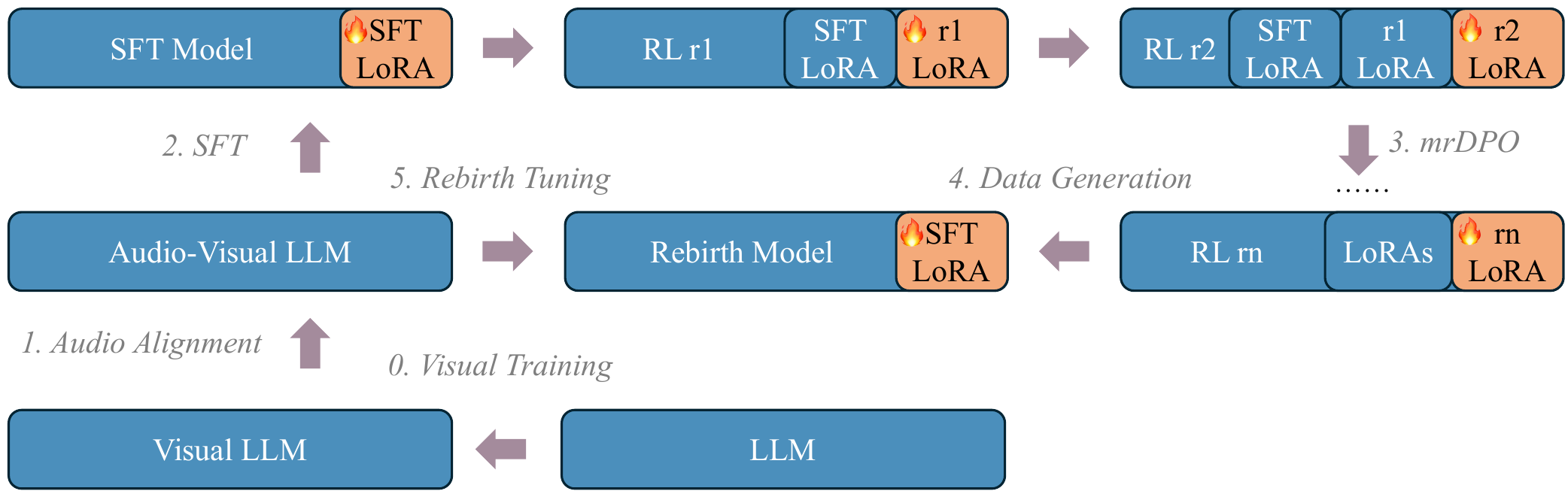}
    \caption{An overview of the complete training process includes audio modality alignment, SFT, mrDPO, and rebirth tuning. LoRA is introduced during the SFT stage, and a new LoRA proxy is added to the LLM before each round of DPO. After multiple rounds of DPO, the model generates new data, which is then utilized during rebirth tuning to further refine the model. }
    \label{fig:rlrebirth}
\end{figure}

Although the model demonstrates the ability to describe synchronized audio-visual information in video after SFT, several issues persist, including missing information, hallucinations, and repetitive decoding. To address these shortcomings, we apply RL based on mrDPO to improve the model's performance. Additionally, we introduce rebirth tuning after RL to further enhance the model's performance in non-captioning tasks. Fig. \ref{fig:rlrebirth} provides an overview of the entire training process involving mrDPO and rebirth tuning, with further details explained in Sections \ref{subsec:mmmrl} and \ref{subsec:rebirth}.

\vspace{-0.3cm}
\subsection{RL stage with mrDPO}
\label{subsec:mmmrl}

We aim to leverage RL to improve the quality of video captions generated by the model. To establish an effective method for evaluating the completeness of video captions, we propose using atomic events as a bridge to automatically assess the preference of caption samples through artificial intelligence (AI) feedback, guiding the model to produce more accurate and detailed video descriptions.


\begin{figure}[ht]
    \centering
    \includegraphics[width=\linewidth]{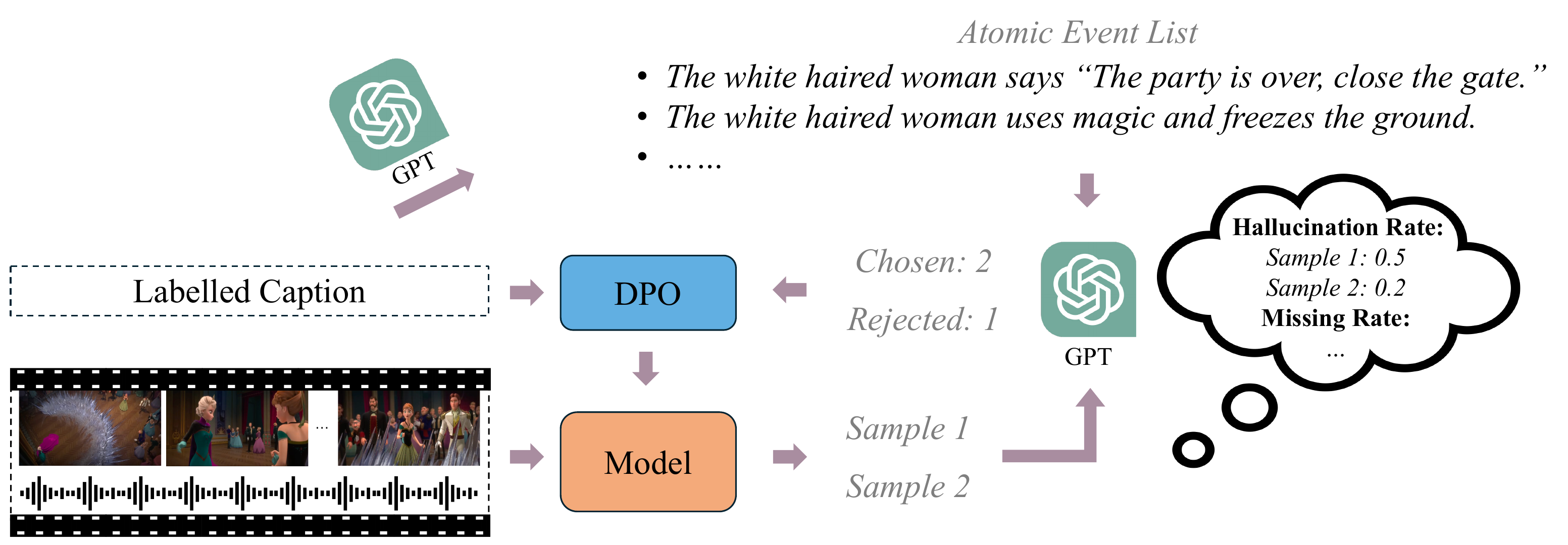}
    \caption{The pipeline for mrDPO in video captioning uses GPT to process the video’s atomic event list to automatically identify preferred and dispreferred caption samples generated by the model. These sample pairs are then utilized in applying DPO-based RL algorithms. This process can be repeated over multiple rounds to iteratively enhance performance.}
    \label{fig:dpo}
\end{figure}

The pipeline for selecting preferred samples when applying RL to video-SALMONN 2 is illustrated in Fig. \ref{fig:dpo}. First, distinct video captions are sampled from the model’s output distribution, given the input video. These captions may be either global captions describing the entire video or local captions focusing on a specific time interval.
To determine the preferred sample for global captions, the labelled caption of the input video is fed into a powerful text LLM, which is tasked with breaking down the caption into basic atomic events. This is relatively straightforward for commercial LLMs like GPT-3.5 and GPT-4o, and the resulting atomic events are generally reasonable.
Next, the text LLM is used to evaluate the caption by identifying missed or hallucinated events, calculating the \textit{information missing and hallucination rates}. The \textit{total error rate} is the sum of the missing and hallucination rates. For local captions, a similar evaluation process is followed, with atomic events being extracted using Gemini-1.5-Pro (as detailed in Appendix~\ref{app:local}).
In addition to metrics based on atomic events, we also consider the \textit{repetition rate} of the video captions. The calculation procedure for this is provided in Appendix \ref{app:rep}.
DPO \citep{rafailov2024direct} is applied as the main RL method based on automatic AI feedback. We assume that only sample pairs with significant metric differences are suitable for RL. Therefore, sample pairs with minimal gaps in metrics are excluded from the RL training set.

Unlike previous approaches that applied only single-round DPO to multimodal LLMs, we introduce a multi-round strategy, as prolonged offline training with a single round fails to optimize the model effectively due to the reference model being biased against the most recent model update in the DPO algorithm. In the multi-round framework, at each $t$th round, the following steps are taken to perform DPO training for the current round.
\begin{enumerate}[itemsep=0pt, leftmargin=*]
\item First, pre-trained LoRA module $\Delta_{t-1}$ is merged into the LLM backbone $\Lambda_{t-1}$ to derive a new LLM backbone $\Lambda_{t}$ that is equivalent to $\Lambda_{t-1}$ with $\Delta_{t-1}$, based on Eqn.~(\ref{eq:merge}): 
\begin{equation}\label{eq:merge}
    \mathbf{W}_{t}=\mathbf{W}_{t-1}+\alpha\mathbf{A}_{t-1}\mathbf{B}_{t-1},
\end{equation}
where $\mathbf{W}_t$ and $\mathbf{W}_{t-1}$ are the weight parameters to adapt in $\Lambda_{t}$ and $\Lambda_{t-1}$, $\alpha$ is the scaling factor of LoRA, 
$r$ is the rank of LoRA, $d$ is the dimension of $\mathbf{W}_{t-1}$. $\mathbf{A}_{t-1}\in \mathcal{R}^{d\times r}$ and $\mathbf{B}_{t-1}\in\mathcal{R}^{r\times d}$ are the low-rank matrix parameters of LoRA in the previous round $t-1$, and $\mathbf{W}\in\mathcal{R}^{d\times d}$ is the parameter of LLM backbone.
\item Next, the new LLM backbone $\Lambda_{t}$ is paired with a new randomly initialized LoRA module $\tilde{\Delta}_{t}$, forming the policy model for round $t$. To address the issue of the increasing difference between the reference and policy models caused by freezing the reference model in standard DPO, $\Lambda_{t}$ is used as the updated reference model used in round $t$. 
\item At last, $\tilde{\Delta}_{t}$ is trained to obtain a well-trained ${\Delta}_{t}$, which can be achieved using the standard DPO loss. However, after multiple training rounds, the model starts to produce unnatural language patterns such as unintelligible or meaningless sentences.
To alleviate 
this issue by stabilizing the training, a guided DPO (gDPO) loss is proposed as    
\begin{align}
    \label{equ:gdpo}
     \mathcal{L}_\text{gDPO}(\mathbf{\pi}_{\theta}; \piref) = \mathcal{L}_\text{DPO}(\mathbf{\pi}_{\theta}; \piref)+\lambda\,\mathbb{E}_{(\mathbf{x}, \mathbf{y}_{\text{gt}})\sim \mathcal{D}_{\text{gt}}}\log \pi_\theta(\mathbf{y}_{\text{gt}}|\mathbf{x}), 
\end{align}
where $\mathcal{L}_\text{DPO}$ is the standard DPO loss, $\mathbf{\pi}_{\theta}=\{\Lambda_{t},\tilde{\Delta}_{t}\}$ and $\piref=\Lambda_{t}$ represent the policy and reference models, respectively. $\mathcal{D}_{\text{gt}}$ denotes the SFT training dataset, where $(\mathbf{x}, \mathbf{y}_{\text{gt}})$ corresponds to a video and its paired ground-truth text description, randomly selected from $\mathcal{D}_{\text{gt}}$. Finally, $\lambda$ is the weight of the second regularization term, which corresponds to cross-entropy learning towards the ground-truth text descriptions without unnatural patterns.
\end{enumerate}
These steps complete the training for a single round. Our proposed mrDPO is implemented by repeating these steps across multiple rounds. Notably, by merging $\Delta_{t-1}$ into $\Lambda_{t-1}$ and equipping the resulting $\Lambda_{t}$ with a new LoRA module $\tilde{\Delta}{t}$, the new $\tilde{\Delta}{t}$ functions as a LoRA proxy for parameter updates. This proxy helps regularize the training by introducing a new random initialization at each round of mrDPO.

\vspace{-0.3cm}
\subsection{Rebirth Tuning}
\label{subsec:rebirth}
After multiple rounds of iteration with the LoRA proxy, the model demonstrates significant improvements in captioning, showcasing the strong potential for audio-visual understanding. However, despite efforts to preserve its language abilities, the model gradually begins to produce repetitive and unnatural text patterns in its responses. In some cases, its benchmark performance remains high even if unnatural patterns appear. 
We believe this issue arises because RL methods primarily optimize the model's output distribution using self-generated data. As a result, the model may overfit the feedback provided by AI models that mismatch with real human preferences, leading it to adopt these unnatural patterns. This tendency can cause a collapse in the model's output distribution, resulting in frequent incoherent or repetitive outputs, and considerable declining performances on the non-captioning tasks.

Rebirth tuning is introduced to address the issue of declining non-captioning language abilities. This method applies teacher-forcing training on self-generated data, promoting a more stable learning process for video understanding. Teacher-forcing, which guides the model to predict the next token, helps prevent it from converging on limited and repetitive patterns. More specifically, before applying rebirth tuning, mrDPO is halted once we observe a significant decline in the model's language capabilities. The final iteration of the model, which excels at generating complete and accurate video descriptions, is then used to label a large dataset of videos. Since the model's language abilities remain relatively intact, natural and fluent descriptions can be easily filtered by detecting problematic patterns, with the remaining high-quality descriptions used as training data for rebirth tuning.

Rebirth tuning is applied to the model after audio modality alignment, allowing it to be "reborn" from self-generated high-quality data to enhance video understanding. Following rebirth tuning, the model not only avoids catastrophic forgetting of non-captioning abilities but also supports the development of the next generation of models by applying mrDPO, followed by the subsequent stage of rebirth tuning.


\vspace{-0.4cm}
\section{Experimental Setup}
\vspace{-0.2cm}
\subsection{Model Specifications}
video-SALMONN 2 is built on an internally trained high-performance visual LLM. This visual LLM uses SigLIP \citep{zhai2023sigmoid} as the visual encoder, Qwen 1.5 with 7B parameters as the backbone LLM, and two linear layers with GELU activation function \citep{hendrycks2016gaussian} as the visual aligner. The model processes video frames at a frame rate of 1 (\textit{i.e.}, $\phi=1$), and can handle up to 30 frames. For videos longer than 30 seconds, 30 frames are uniformly sampled from the video.

For the audio branch, we use the Whisper-Large-v3 encoder \citep{radford2023robust} as the audio encoder, and a window-level Q-Former \citep{tang2024extending} with a window length of 0.2 seconds as the audio aligner, producing a total of 150 audio tokens for a 30-second input. The Whisper encoder has a maximum processing duration of $t_{\text{max}} = 30$ seconds.
The rank $r$ and scaling factor $\alpha$ of LoRA are set to 256 and 2.0, respectively. During training, the visual encoder, visual aligner, audio encoder, and LLM remain frozen.





\vspace{-0.3cm}
\subsection{Data and Training Specifications}
During the audio modality alignment stage, LibriSpeech-960h \citep{panayotov2015librispeech} and AudioCaps \citep{audiocaps} are used to train the audio aligner. LibriSpeech-960h is utilized for speech recognition training, while AudioCaps is employed for audio captioning training.

In the audio-visual SFT stage, experiments are conducted using an internal video dataset that will be released upon acceptance. A total of 13k videos with rich audio information are automatically selected, and high-quality audio-visual captions are regenerated with the assistance of GPT-4o \citep{openai2024gpt4technicalreport}, Whisper-Large-v3 \citep{radford2023robust}, and SALMONN \citep{tang2024salmonn}. The detailed pipeline is described in Appendix~\ref{app:sft_data}. Additionally, to further enhance the quality of the SFT data, around 1.5k video captions were manually refined.

In the RL stage, two kinds of tasks are studied: global captioning for the whole video and local captioning for a given time interval. Before each round, a pair of captions for both global and local captioning are sampled from the model for each video in SFT data, respectively.
We consider the information missing rate, hallucination rate, and repetition rate to determine whether a sample pair is suitable for DPO and determine the chosen and rejected samples if yes. The selecting methods for each round are listed in Appendix~\ref{app:select}.

After mrDPO, the language abilities of the model are reduced. We stop RL training once significant degradation in language abilities is detected. The checkpoint of the last DPO round is used to label captions of a large number of videos. Unnatural captions are eliminated, and the remaining high-quality captions form the training data for the rebirth tuning stage.

For the test dataset, we curated a video captioning benchmark to evaluate the event missing rate (Miss), hallucination rate (Hall), and text repetition rate (Rep). Details of the test data and evaluation process can be found in Appendix~\ref{app:testset} and Appendix~\ref{app:eval}, respectively. The benchmark consists of 483 carefully selected videos, each labelled with complete audio-visual captions by human annotators. Atomic events for the test dataset were initially obtained using GPT-4o and then manually refined. For local captioning, we used Gemini-1.5-Pro \citep{geminiteam2024geminifamilyhighlycapable} to tag the start and end times of each event within specific time intervals. Since Gemini could not process some videos, only 457 videos were used for the local captioning evaluation.

Regarding training settings, we conducted audio modality alignment using 8$\times$A100 GPUs for 35k steps and audio-visual SFT using 16$\times$A100 GPUs for 4 epochs. Each RL round was trained with 64$\times$A100 GPUs for 1k steps. After six rounds of mrDPO training, rebirth tuning was performed. During the rebirth tuning stage, we used 64$\times$A100 GPUs and trained for 4 epochs. The batch size per device was set to 1, making the total batch size equal to the number of devices. The weight $\lambda$ in Eqn.~(\ref{equ:gdpo}) was set to 0.1 for all related experiments. The final video-SALMONN 2 model was obtained after one round of gDPO training following rebirth tuning.

\vspace{-0.4cm}
\section{Experimental Results}
\vspace{-0.2cm}
\subsection{Overall Results}
The results of our video captioning benchmarks are presented in Table \ref{tab:res_I}. video-SALMONN 2 outperforms other models in both information missing and hallucination rates for global and local captioning. Among existing open-source multimodal LLMs, few can provide detailed and accurate video descriptions, whether purely visual models like Video-LLaVA \citep{lin2023video} and VILA \citep{lin2023vila}, or audio-visual models like Video-LLaMA 2 \citep{damonlpsg2024videollama2} and video-SALMONN \citep{sun2024videosalmonn}. Notably, many open-source models, such as Video-LLaVA and Video-LLaMA 2, tend to generate shorter captions, leading to high information missing rates but low hallucination rates.
GPT-4o and Gemini-1.5-Pro can generate more detailed captions and are of higher quality than current open-source models. However, the purely visual version of GPT-4o lacks audio comprehension, and Gemini's understanding of visual content is somewhat limited, resulting in both models exhibiting some degree of information missing and hallucination.

Our visual base model, trained on a large dataset of images and silent videos, is capable of generating detailed text descriptions based solely on visual information, with a relatively low information missing rate. However, generating longer texts leads to a higher hallucination rate. After audio modality alignment and audio-visual SFT, the model can leverage audio content to reduce both information loss and hallucinations in its descriptions. However, the inclusion of audio tokens may confuse the visual LLM, resulting in frequent repetition during decoding. Building on the SFT model, we applied mrDPT and rebirth tuning, achieving approximately a 35\% absolute reduction in the repetition rate and absolute reductions of around 40\% and 20\% in the total error rate for global and local captioning, respectively. The final video-SALMONN 2 model outperforms some commercial models like GPT-4o and Gemini-1.5-pro in video captioning. As an audio-visual LLM, video-SALMONN 2 retains strong visual understanding capabilities and performs well on various visual benchmarks, such as Video-MME \citep{fu2024video}. For more details, refer to Appendix~\ref{app:vres}.

\begin{table}[ht]
    \caption{Results of our benchmark for detailed video captioning evaluation. ``A'' and ``V'' refer to the audio and visual modalities respectively. The repetition rate (Rep), event missing rate (Miss), hallucination rate (Hall), and total error rate (Total = Miss + Hall) are assessed for the captions. video-SALMONN 2, which undergoes an additional round of gDPO after rebirth tuning, achieves the best performance in both global and local captioning, with the lowest total error rates.}
    \setlength{\tabcolsep}{3pt}
    \small
    \centering
      \begin{tabular}{lcccccccc}
      \toprule
      \multirow{2}{*}{\textbf{Model}} & \multirow{2}{*}{\textbf{Modality}} & \multicolumn{4}{c}{\textbf{Global}} & \multicolumn{3}{c}{\textbf{Local}}  \\
      \cmidrule(lr){3-6} \cmidrule(lr){7-9}
      & & \textbf{\%Rep$\downarrow$} & \textbf{\%Miss$\downarrow$} & \textbf{\%Hall$\downarrow$} & \textbf{\%Total$\downarrow$} & \textbf{\%Miss$\downarrow$}  & \textbf{\%Hall$\downarrow$} & \textbf{\%Total$\downarrow$} \\
      \midrule
       GPT-4o Visual & V & 3.6 & 16.6 & 17.2 & 33.8 & 35.3 & 30.7 & 66.0  \\
       Gemini-1.5-Pro & A + V & 1.3 &  21.8 & 16.5 & 38.3 & 36.9 & 17.2 & 54.1 \\
      \midrule
      7B Video-LLaVA & V & 13.2 & 65.3 & \textbf{5.4} & 70.7 & 59.1 & \textbf{9.4} & 68.5 \\
      8B VILA & V & 4.5 & 39.3 & 18.6 & 57.9 & 47.9 & 23.4 & 71.2 \\
      7B Video-LLaMA 2 & A + V & 5.7 & 56.8 & 8.9 & 65.7 & 47.6 & 14.3 & 61.9 \\
      13B video-SALMONN & A + V & \textbf{1.2} & 52.1 & 26.6 & 78.7 & 47.8 & 40.7 & 88.4 \\
      \midrule
      7B Ours-Visual Base & V & 11.8 & 29.8 & 30.0 & 59.7 & 36.1 & 46.1 & 82.2 \\
      7B Ours-SFT & A + V & 36.0 & 26.7 & 26.9 & 53.6 & 30.8 & 33.3 & 64.0 \\
      7B video-SALMONN 2 & A + V & 1.4 & \textbf{6.9} & 6.8 & \textbf{13.7} & \textbf{22.2} & 21.4 & \textbf{43.6}   \\
      \bottomrule
      \end{tabular}
  \label{tab:res_I}
\end{table}

\subsection{Analysis of Multi-Round Reinforcement Learning}
\label{ssec:unnatural}

\begin{figure}
    \centering
    \begin{subfigure}{0.45\linewidth}
        \centering
        \includegraphics[width=\linewidth]{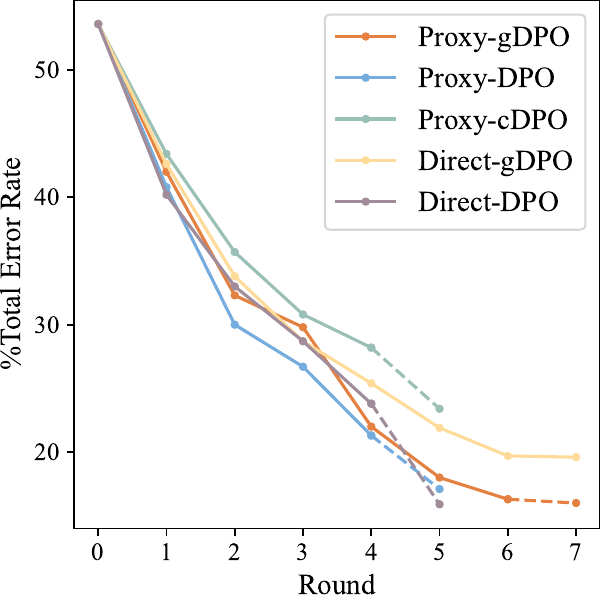}
        \caption{Comparison of different mrDPO settings.}
        \label{fig:loss}
    \end{subfigure}
    \begin{subfigure}{0.53\linewidth}
        \centering
        \includegraphics[width=\linewidth]{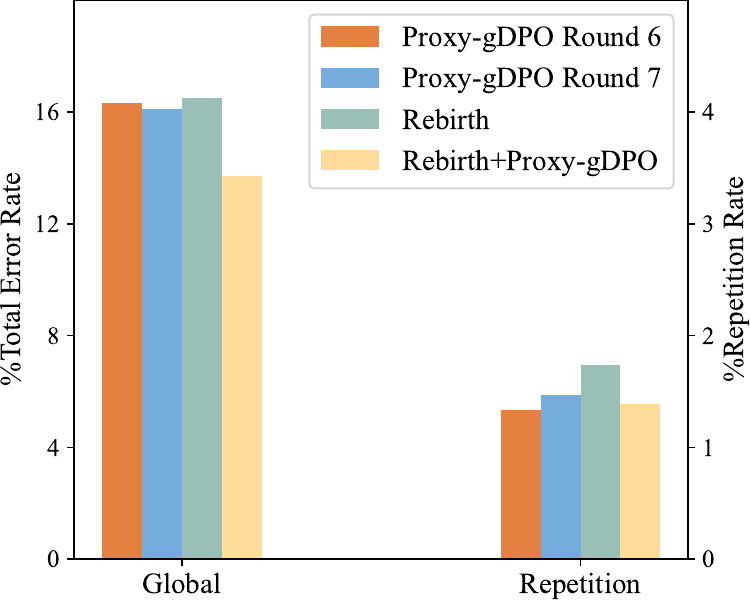}
        \vspace*{-0.02cm}
        \caption{gDPO with LoRA proxy vs. Rebirth tuning.}
        \label{fig:rl}
    \end{subfigure}
    \caption{The comparison of different methods is shown in Fig. \ref{fig:loss}. All models are trained either until convergence or until they begin to frequently generate unnatural patterns. The dashed line indicates that this round of RL resulted in a high frequency of unnatural text generation and, therefore, cannot be considered a valid performance indicator for the model. Fig. \ref{fig:rl} demonstrates that RL after rebirth tuning achieves better performance, while mrDPO provides only minimal gains.}
\end{figure}

This section explores various approaches to training in mrDPO. In terms of loss functions, we compared the standard DPO loss, the proposed gDPO loss (which includes the regularisation term based on the ground-truth captions), and a similar loss referred to as ``cDPO", which is the sum of DPO loss and cross-entropy loss on chosen samples. Additionally, the LoRA proxy is evaluated against directly tuning the model's original LoRA, referred to as ``Direct". Fig.~\ref{fig:loss} presents the total error rates for global video captioning. 
Training is halted when unnatural captions begin to appear frequently. Examples of such cases are provided in Appendix~\ref{app:badcases}.

Among the three loss functions, DPO shows the fastest improvement in captioning metrics, but unfortunately, it also quickly leads to outputs with frequent unnatural patterns. This is likely because the model only sees self-generated labels rather than ground-truth labels. cDPO faces the same issue but performs worse than DPO. By incorporating loss on ground-truth labels, gDPO makes the training process more stable, allowing the model to generate text responses without unnatural patterns for a longer period of training across multiple RL rounds. This stability also preserves the model's potential for further improvement, with a significant drop in the error rate observed after three rounds of mrDPO.

The LoRA proxy that randomly initializes a new LoRA module in each RL round, is found to be more beneficial for mrDPO performance compared to directly training the same LoRA, especially after over 4 gDPO rounds. This reveals that the regularisation effect introduced by the LoRA proxies helps to alleviate over-fitting. 
Since gDPO with LoRA proxy performs the best for mrDPO, we use the model after six gDPO rounds using LoRA proxy to generate captions for a large number of videos. After excluding unnatural patterns, a total of 180k video captions remain for rebirth tuning.

\subsection{Analysis of Rebirth Tuning}

\begin{table}[ht]
    \caption{The appearance rate of unnatural caption for global. We detect specific patterns that are viewed as unnatural and count the frequency of occurrence of specific patterns on the test set over mrDPO using gDPO with LoRA proxy. The results prove that mrDPO leads to a significant increase in the appearance of unnatural captions, especially in the final converging rounds. }
    \setlength{\tabcolsep}{4pt}
    \centering
      \begin{tabular}{lccccccccc}
      \toprule
      \multirow{2}{*}{\textbf{Stage}} & \multirow{2}{*}{\textbf{SFT}} & \multicolumn{7}{c}{\textbf{\#Rounds of mrDPO}} & \multirow{2}{*}{\textbf{Rebirth Tuning}} \\
      \cmidrule(lr){3-9}
      & & 1 & 2 & 3 & 4 & 5 & 6 & 7 & \\
      \midrule 
      \textbf{\%Unnatural Rate$\downarrow$} & ~~0.0~ & 0.0 & 0.0 & 1.9 & 0.9 & 2.9 & 2.1 & 12.0 & 0.0 \\
      \bottomrule
      \end{tabular}
  \label{tab:pattern}
\end{table}



While multiple rounds of mrDPO with LoRA proxy significantly improve video captioning performance, they also lead to an increasing frequency of unnatural patterns in text responses. Table \ref{tab:pattern} shows the occurrence rate of these unnatural patterns in global captioning after each training stage. Through rebirth tuning, the backbone LLM discards the LoRA proxies and restores its ability to generate fluent captions. Additionally, the careful selection of rebirth-tuning data enhances data quality, ensuring the model is fine-tuned with superior data, further boosting its overall performance.

Another notable effect of rebirth tuning is to sustain continued training. As shown in Fig.~\ref{fig:loss}, in the later rounds of mrDPO, the model starts to gain less with each round and eventually converges. The decline in the ability to generate fluent text responses is also more likely to occur in these later rounds, suggesting that the model has fallen into a local minimum after mrDPO.
However, after the rebirth tuning stage, where only teacher-forcing training is applied, the model escapes the local optimum from previous training and becomes receptive to further optimization with RL. Fig.~\ref{fig:rl} compares gDPO after rebirth tuning with six rounds of gDPO with LoRA proxy. It is observed that only minimal improvement can be achieved after sufficient RL rounds in terms of the total error rate for global captioning, while an extra RL stage following rebirth tuning yields significant performance gains once again.
This suggests the potential of iterating mrDPO and rebirth tuning. 

\section{Conclusions}
This work introduces video-SALMONN 2, a powerful audio-visual LLM designed for detailed video captioning, and proposes the mrDPO method. To our knowledge, this is the first study of applying RL to audio-visual LLMs in literature. New metrics are designed to evaluate the information missing and hallucination rates in video captions, which are used to guide sample selection for DPO. To further stabilize training, the setting with novel gDPO and LoRA proxy is introduced. After mrDPO, we propose a novel rebirth tuning method to restore LLM's performance on non-captioning tasks. As a result, video-SALMONN 2 demonstrates significant improvements in video captioning, outperforming notable models such as GPT-4o and Gemini-1.5-Pro, and setting a promising direction for achieving detailed and accurate video captioning for video understanding.

\bibliography{iclr2025_conference}
\bibliographystyle{iclr2025_conference}

\newpage
\appendix

\section{Pipeline for Getting Atomic Events in a Time Interval}
\label{app:local}
Gemini-1.5-Pro is used to obtain the atomic events in some time intervals. We input the video and all its atomic events labelled by GPT to Gemini and ask it to tag the beginning and start time of each atomic event. Events are then selected if their time intervals overlap with the given time interval.
We have checked this process to confirm that the atomic events obtained for the given time interval are roughly accurate.

\section{Calculation Procedure of The Text Repetition Rate}
\label{app:rep}
The procedure to calculate the repetition rate of a long and detailed text is shown as follows:
\begin{enumerate}
    \item Split the text into short phrases by punctuation;
    \item Counting the number of occurrences of each phrase;
    \item The number of recurring phrase words divided by the total number of words in the text is the repetition rate.
\end{enumerate}


\section{Pipeline for Labelling High-quality Audio-visual Captions}
\label{app:sft_data}
\begin{figure}[h]
    \centering
    \includegraphics[width=\linewidth]{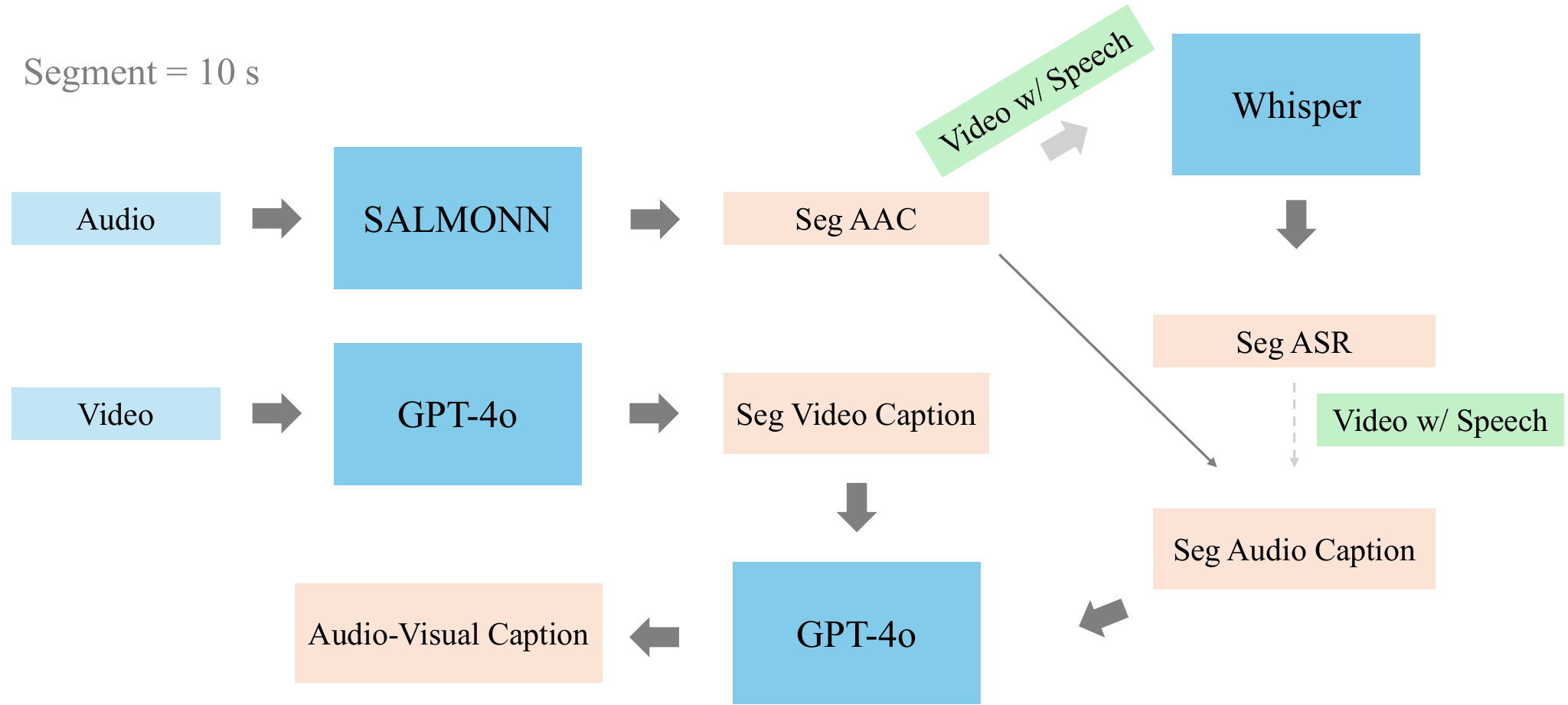}
    \caption{The pipeline for labelling videos with high-quality audio-visual captions.}
    \label{fig:data_pipeline}
\end{figure}

To curate training data for supervised fine-tuning, we employ GPT-4o to label the visual content in each frame, while SALMONN-13B and Whisper-Large-v3 are used to annotate the speech content and audio events in the audio track. This process is illustrated in Figure~\ref{fig:data_pipeline}. Our initial aim is to automatically filter out videos that contain limited speech. We begin by slicing each video into 10-second segments, with the audio from each segment analyzed by SALMONN to generate automatic audio captions (AAC). These captions help us filter out videos that lack descriptive speech such as "A man is speaking" or "A woman says...". This initial filtering step is somewhat coarse.

Next, the audio from each segment is processed by Whisper to produce automatic speech recognition (ASR) results. If the transcribed text is too brief or nonsensical, the corresponding video is deemed to lack rich audio content and is excluded from further consideration. For a video to be labelled, all of its segments should pass this exclusion criterion.

The segments from the remaining videos are then sampled at a rate of 1 fps and fed into GPT-4o to extract segment-level visual captions. Ultimately, the segment-level visual captions, AAC results, and ASR results form the input to GPT-4o concurrently to generate a detailed global audio-visual caption for each video.

\section{About The Test Dataset}
\label{app:testset}

\begin{figure}
    \centering
    \begin{subfigure}{0.495\linewidth}
        \centering
        \includegraphics[width=\linewidth]{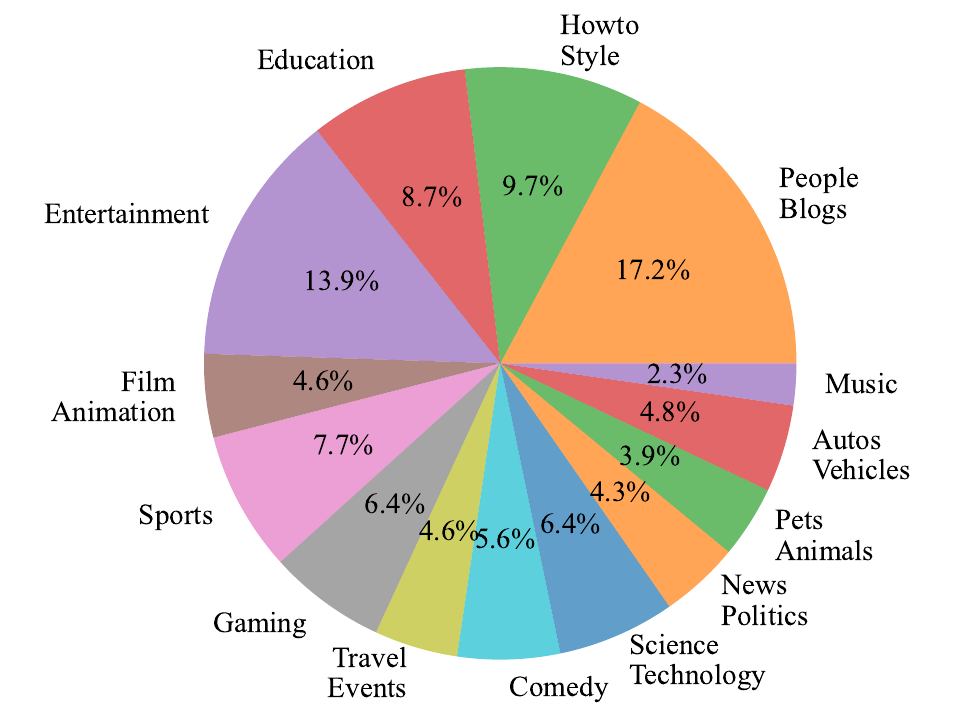}
        \caption{Video type distribution.}
    \end{subfigure}
    \begin{subfigure}{0.495\linewidth}
        \centering
        \includegraphics[width=\linewidth]{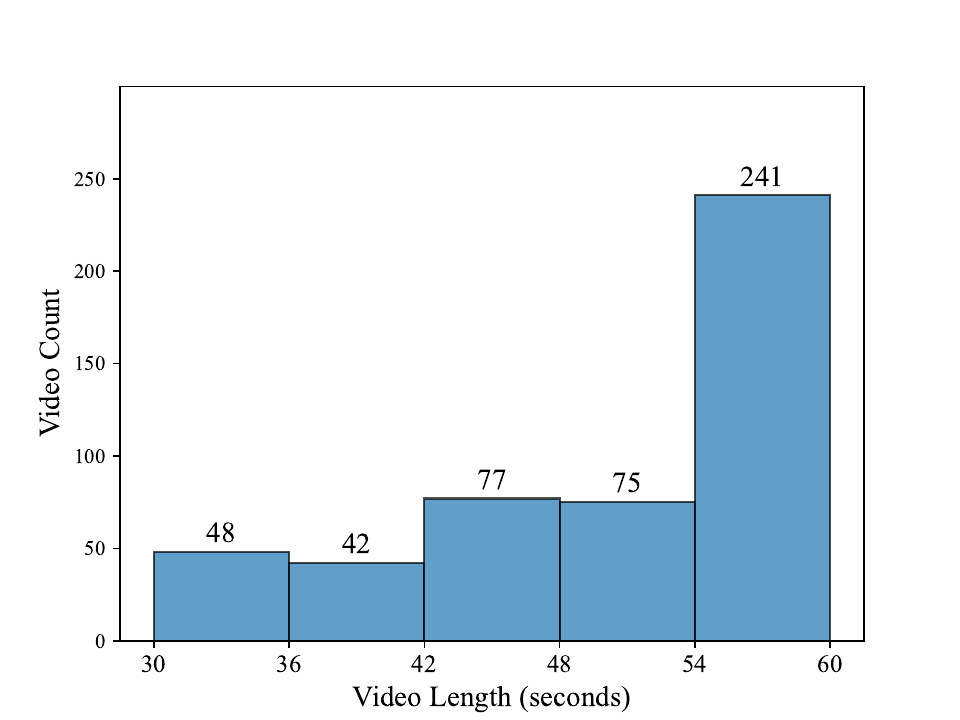}
        \caption{Video length distribution.}
    \end{subfigure}
    \caption{The basic information of our benchmark data. }
    \label{fig:benchmark}
\end{figure}

Figure \ref{fig:benchmark} shows the basic information of our caption benchmark. The benchmark covers 14 different fields. All the videos are between 30s to 60s, with an average duration of 51s.

\section{Process of Evaluating Detailed Captions}
\label{app:eval}
To evaluate the specific video caption generated by our model, we first use GPT-3.5 or GPT-4o to split the labelled caption of this video into several atomic events, where we use GPT-4o for the test set and GPT-3.5 for the RL training set. Then, the list of atomic events and the caption to be evaluated are simultaneously fed into GPT-3.5 to determine what events in the atomic event list are missed and what events in the caption are hallucinated.
Specifically, we ask GPT-3.5 to list all the missing events and hallucination events for better evaluation precision. Note that events that are described incorrectly are also regarded as hallucinations. 
The quotient between the number of missing or hallucination events and the number of all events in the video is the final missing or hallucination rate.
For more robust testing, GPT-3.5 is used to evaluate 7 times for each caption and the medium number of the metric is reported. We have manually confirmed that the score GPT-3.5 gives is roughly plausible.

\section{Samples Selecting Methods for RL of Each Round}
\label{app:select}
To achieve better performance and training efficiency, we take a specially designed strategy to select proper preference pairs. A sample pair is selected if one sample is better than the other in all metrics with a threshold. For global captioning, we consider global error rate $\Delta e_g$ and global repetition rate $\Delta r_g$, while for local captioning we consider local error rate $\Delta e_t$ and local repetition rate $\Delta r_t$. Table \ref{tab:sampleselect} shows the threshold used in each round.

\begin{table}[ht]
    \caption{The data selecting threshold used in each RL round. A negative number means that the chosen sample can be worse than the rejected sample in this metric to some degree. }
    \setlength{\tabcolsep}{4pt}
    \centering
      \begin{tabular}{ccccc}
      \toprule
      \multirow{2}{*}{\textbf{RL Round}} & \multicolumn{4}{c}{\textbf{Threshold Used}} \\
      \cmidrule(lr){2-5}
      & $\Delta e_g$ & $\Delta r_g$ & $\Delta e_t$ & $\Delta r_t$ \\
      \midrule 
      1 & $\ge5\%$ & $\ge1\%$ & $\ge20\%$ & $\ge1\%$\\
      2 & $\ge20\%$ & $\ge-1\%$ & $\ge45\%$ & $\ge0$ \\
      3 & $\ge20\%$ & $\ge-1\%$ & $\ge45\%$ & $\ge0$ \\
      4 & $\ge20\%$ & $\ge-1\%$ & $\ge45\%$ & $\ge0$ \\
      5 & $\ge20\%$ & $\ge-1\%$ & $\ge45\%$ & $\ge0$ \\
      6 & $\ge25\%$ & $\ge-1\%$ & $\ge45\%$ & $\ge0$ \\
      7 & $\ge30\%$ & $\ge-1\%$ & $\ge45\%$ & $\ge0$ \\
      \bottomrule
      \end{tabular}
  \label{tab:sampleselect}
\end{table}

\section{Results on Visual QA Benchmarks}
\label{app:vres}
Since QA data is not seen during the mrDPO process, the model's QA abilities decrease a lot after mrDPO. Thanks to the rebirth tuning on captioning and QA, the non-captioning abilities are able to recover.
After one round of gDPO with a LoRA proxy, video-SALMONN 2 finally achieves detailed and accurate captioning while getting competitive results compared to SOTA models of similar size on QA benchmarks like Video-MME.
Since video-SALMONN 2 cannot support very long audio, we only consider the Video-MME Short set. Table~\ref{tab:vres} shows the results of our models, as well as those of SOTA models of 7B and 8B.


\begin{center}
\begin{threeparttable}[ht]
    \caption{Accuracy results on Video-MME Short set.}
    \setlength{\tabcolsep}{4pt}
    \centering
      \begin{tabular}{lcc}
      \toprule
      \textbf{(\#Params) Model} & \textbf{\#Max Frames} & \textbf{Video-MME Short$\uparrow$} \\ 
      \midrule
      (8B) MiniCPM-V 2.6 & 64 & ~71.3 \tnote{1} \\ 
      (7B) Long-LLaVA & 64 & ~61.9 \tnote{1} \\
      \midrule
      (7B) Ours-Visual Base & 30 & 67.2 \\ 
      (7B) Ours-mrDPO & 30 & 65.3 \\
      (7B) Ours-Rebirth & 30 & 67.6 \\
      (7B) video-SALMONN 2 & 30 & 67.0 \\ 
      \bottomrule
      \end{tabular}
  \label{tab:vres}
  \begin{tablenotes}
        \small
        \item[1] \href{https://video-mme.github.io/home_page.html#leaderboard}{https://video-mme.github.io/home\_page.html\#leaderboard}.
    \end{tablenotes}
\end{threeparttable}
\end{center}


\section{Video Captioning Cases of video-SALMONN 2}
\label{app:cases}
\begin{figure}[ht]
    \centering
    \includegraphics[width=\linewidth]{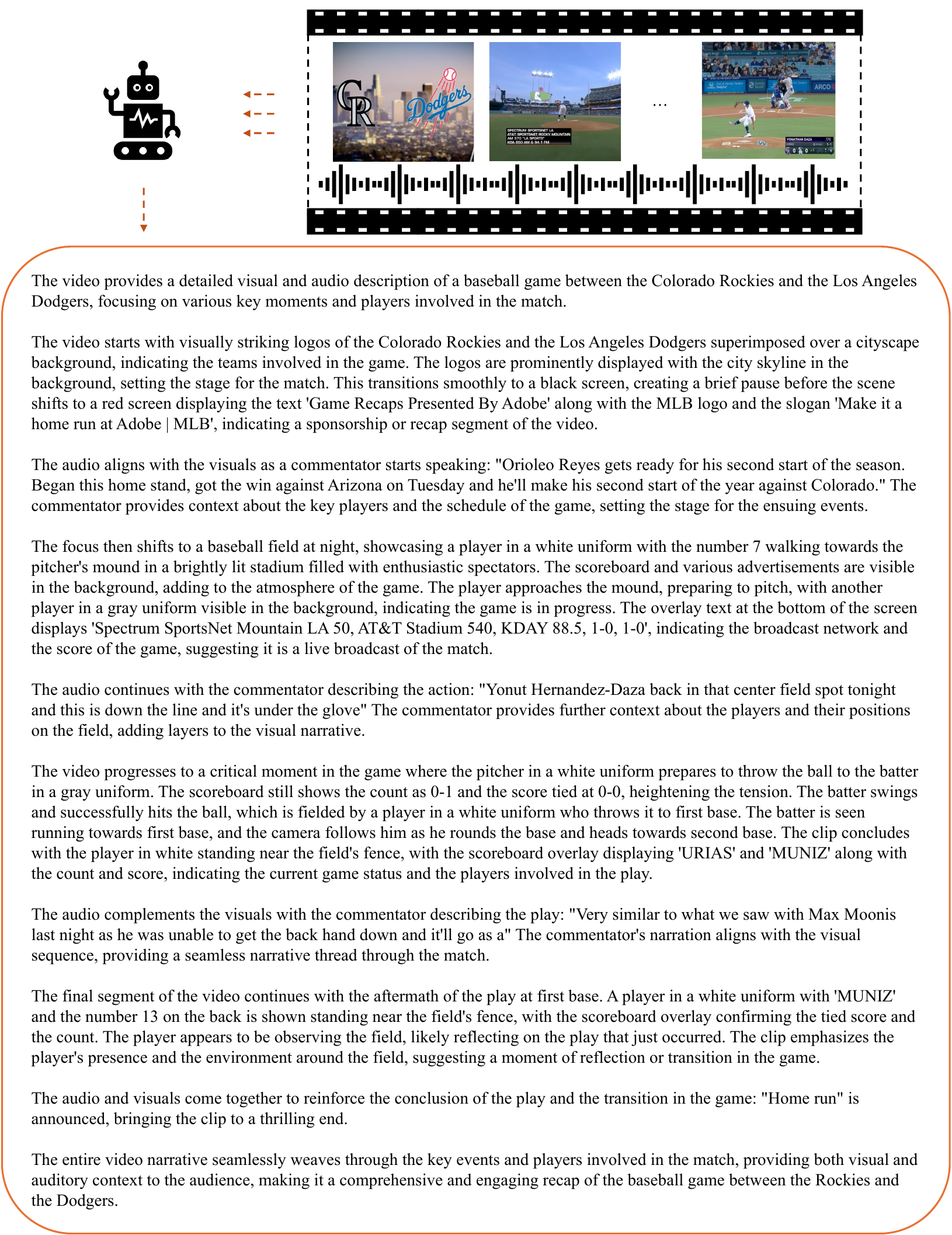}
    \caption{Case I generated by video-SALMONN 2.}
    \label{fig:baseball}
\end{figure}

\begin{figure}[ht]
    \centering
    \includegraphics[width=\linewidth]{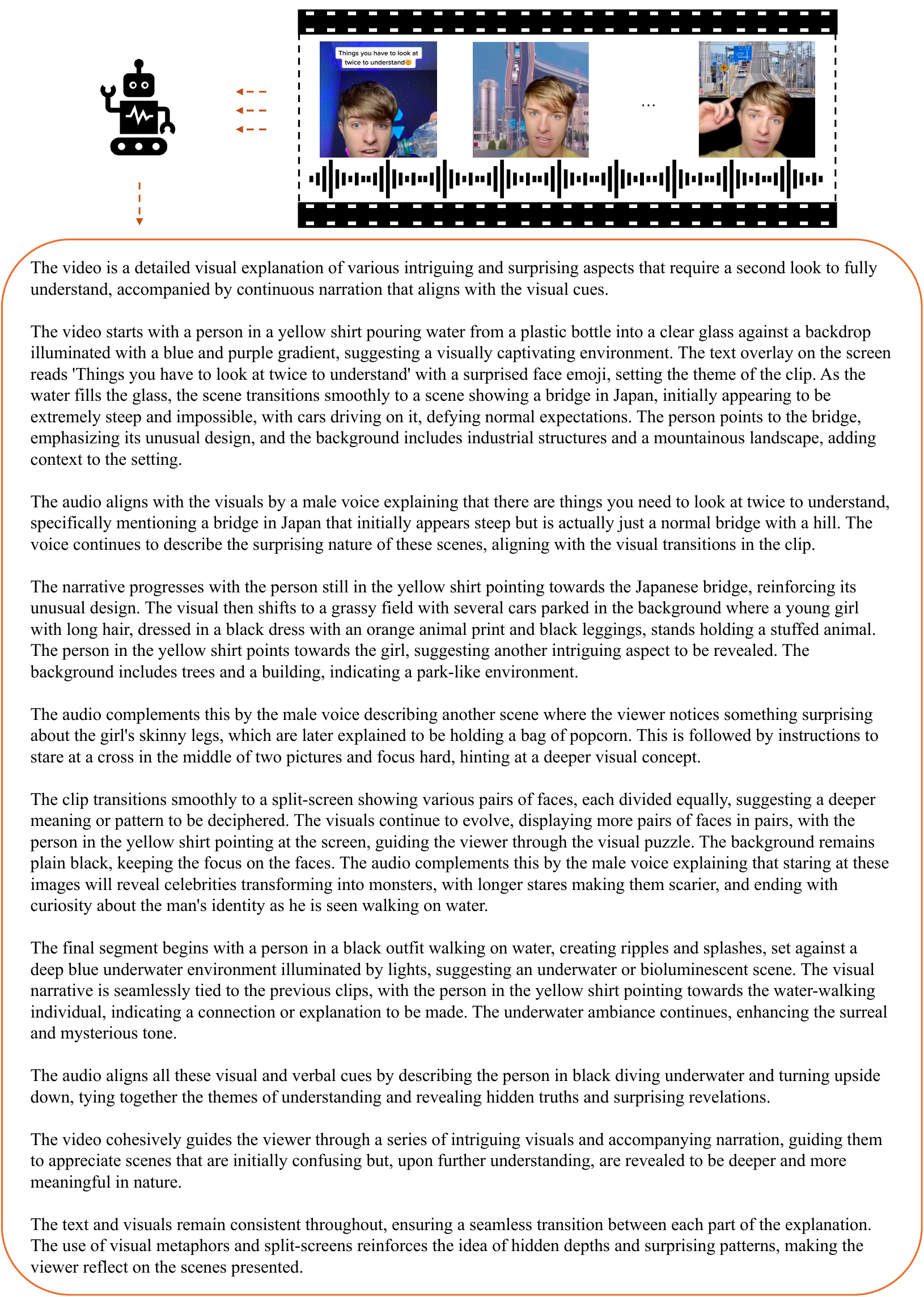}
    \caption{Case II generated by video-SALMONN 2.}
    \label{fig:badcase_good}
\end{figure}

Some video captioning cases generated by video-SALMONN 2 are shown in Figure~\ref{fig:baseball} and Figure~\ref{fig:badcase_good}. More demos can be found at \href{https://video-salmonn-2.github.io}{https://video-salmonn-2.github.io}.

\section{Cases of Unnatural Responses}
\label{app:badcases}

\begin{figure}[ht]
    \centering
    \includegraphics[width=\linewidth]{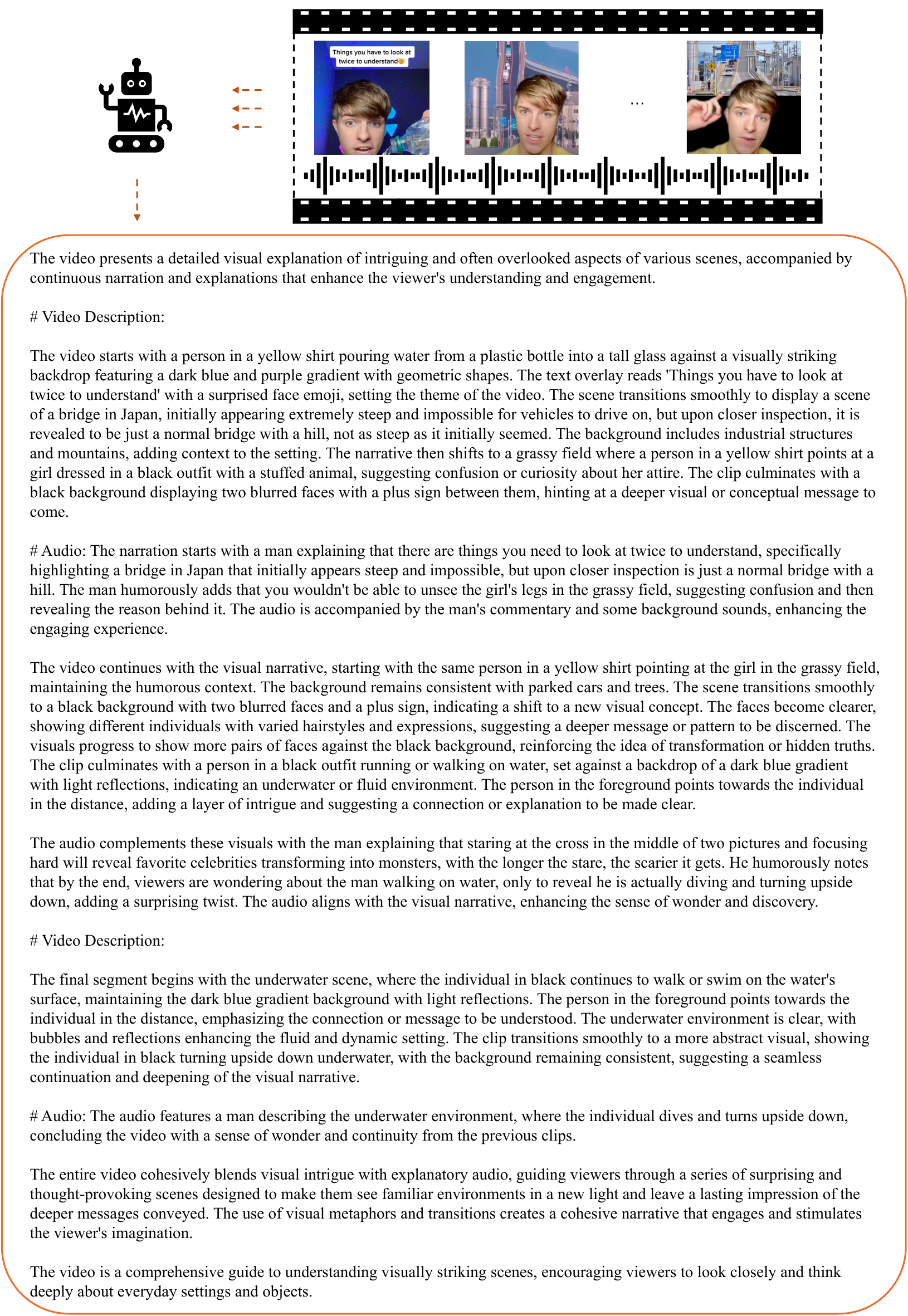}
    \caption{A bad caption generated by the model after 6 gDPO rounds with LoRA proxies, which contains a lot of patterns like ``\#Video Description" and ``\#Audio:" that make the caption unnatural. }
    \label{fig:badcase_mid}
\end{figure}
\begin{figure}[ht]
    \centering
    \includegraphics[width=\linewidth]{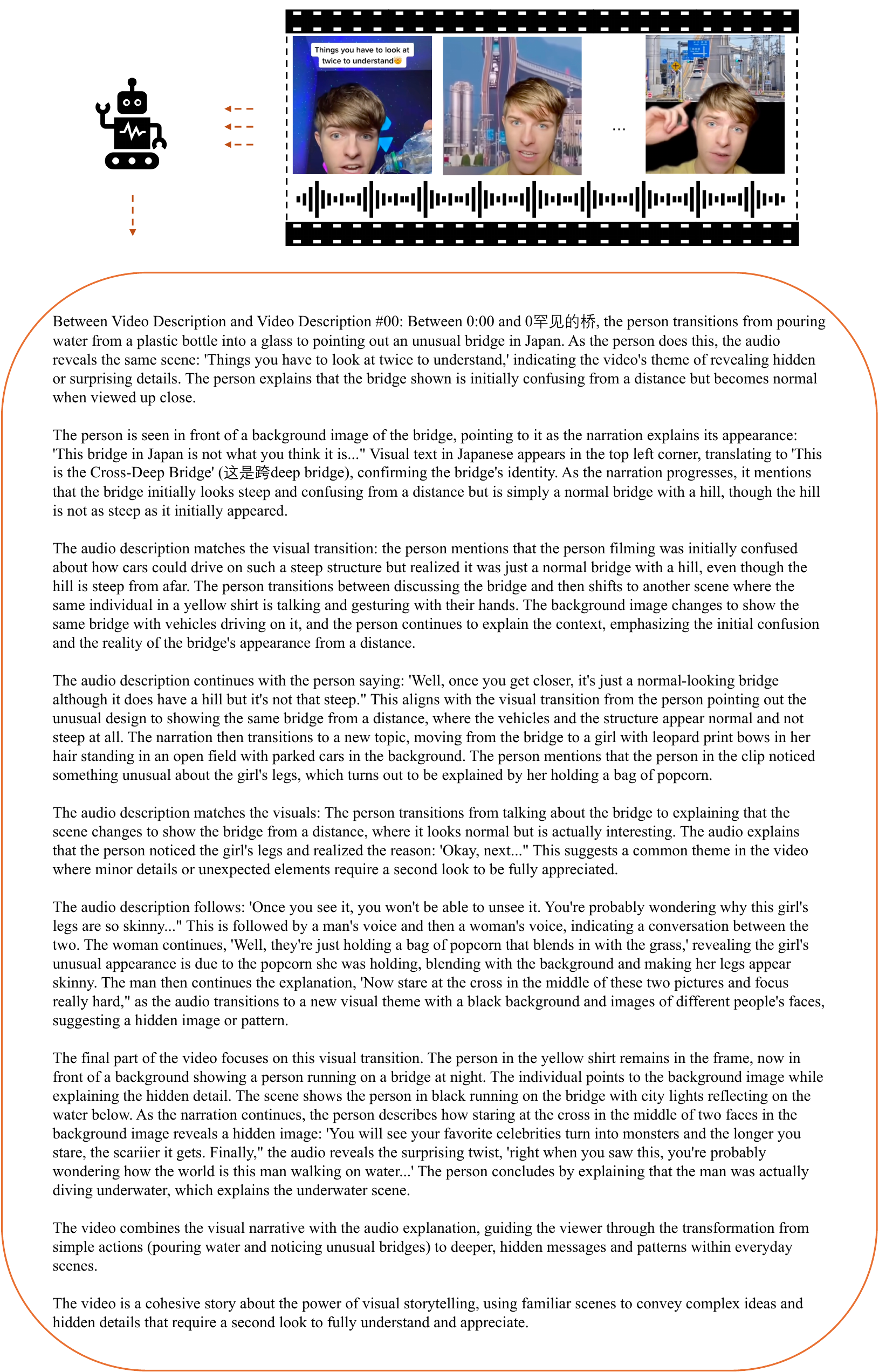}
    \caption{A bad caption generated by the model after 5 classical DPO rounds with LoRA proxies, which includes strange characters and sentences, like the first sentence in this generated text.
}
    \label{fig:badcase_bad}
\end{figure}

Unnatural responses might be generated after multiple RL rounds.
Figure \ref{fig:badcase_mid} shows a caption generated by the model after 6 gDPO rounds with LoRA proxies, which contains some repeated patterns that make the caption unnatural. Figure \ref{fig:badcase_bad} shows a caption generated by the model after 5 classical DPO rounds with LoRA proxies, which includes some strange characters and sentences.

\end{document}